\documentclass[11pt,a4paper]{article}
\usepackage{acl2015}
\usepackage{times}
\usepackage{url}
\usepackage{latexsym}

\usepackage{graphicx}
\usepackage{balance}
\usepackage{url}
\usepackage{xspace}
\usepackage{algorithmic}
\usepackage{algorithm}
\usepackage{multirow}

\usepackage{epsfig}
\usepackage{amssymb}
\usepackage{amsmath}
\usepackage{amsfonts}

\usepackage{placeins}
\usepackage{multirow}

\usepackage{soul}

\def\ie{\emph{i.e.,~}}
\def\eg{\emph{e.g.,~}}

\DeclareMathOperator*{\argmax}{\arg\!\max}

\newtheorem{defn}{Definition}

\newcounter{Lcount}
\newcommand{\numsquishlist}{
  \begin{list}{\arabic{Lcount}. }
   { \usecounter{Lcount}
 \setlength{\itemsep}{0pt}      \setlength{\parsep}{3pt}
     \setlength{\topsep}{3pt}       \setlength{\partopsep}{0pt}
     \setlength{\leftmargin}{2.7em} \setlength{\labelwidth}{1em}
     \setlength{\labelsep}{0.5em} } }
\newcommand{\numsquishend}{\end{list}}

\newcommand{\squishlist}{
  \begin{list}{$\bullet$}
   { \setlength{\itemsep}{0pt}      \setlength{\parsep}{3pt}
     \setlength{\topsep}{3pt}       \setlength{\partopsep}{0pt}
     \setlength{\leftmargin}{1.5em} \setlength{\labelwidth}{1em}
     \setlength{\labelsep}{0.5em} } }
\newcommand{\squishend}{\end{list}}

\makeatletter
\renewcommand\normalsize{%
\@setfontsize\normalsize{10}{12.5}}
\makeatother
\normalsize

\title{A Framework for Comparing Groups of Documents}

\author{Arun S. Maiya \\
  Institute for Defense Analyses --- Alexandria, VA, USA \\
  {\tt amaiya@ida.org}} 

\date{}

\begin{document}
\maketitle
\begin{abstract}
  We present a general framework for comparing multiple groups of documents. A bipartite graph model is proposed where document groups are represented as one node set and the comparison criteria are represented as the other node set.  Using this model, we present basic algorithms to extract insights into similarities and differences among the document groups.  Finally, we demonstrate the versatility of our framework through an analysis of NSF funding programs for basic research.
\end{abstract}

\section{Introduction and Motivation}
\label{sec:intro}

Given multiple sets (or groups) of documents, it is often necessary to {\em compare} the groups to identify similarities and differences along different dimensions.  In this work, we present a general framework to perform such comparisons for extraction of important insights.  Indeed, many real-world tasks can be framed as a problem of comparing two or more {\em groups} of documents.  Here, we provide two motivating examples. 

~\\
\noindent
{\bf 1. Program Reviews.} To better direct research efforts, funding organizations such as the National Science Foundation (NSF), the National Institutes of Health (NIH),  and the Department of Defense (DoD), are often in the position of reviewing research programs via their artifacts (\eg grant abstracts, published papers, and other research descriptions).  Such reviews might involve identifying overlaps across different programs, which may indicate a duplication of effort.  It may also involve the identification of unique, emerging, or diminishing topics.  A ``document group'' here could be defined either as a particular research program that funds many organizations, the totality of funded research conducted by a specific organization, or all research associated with a particular time period (\eg fiscal year).  In all cases, the objective is to draw comparisons {\em between} groups by comparing the document sets associated with them.

~\\
\noindent
{\bf 2. Intelligence.}  In the areas of defense and intelligence, document sets are sometimes obtained from different sources or entities.  For instance, the U.S. Armed Forces sometimes seize documents during raids of terrorist strongholds.\footnote{See {\em Document Exploitation (DOCEX)} at \\\url{http://en.wikipedia.org} for more information.} Similarities between two document sets (each captured from a different source) can potentially be used to infer a non-obvious association between the sources.

~\\
Of course, there are numerous additional examples across many domains (\eg comparing different news sources, comparing the reviews for several products, etc.).  Given the abundance of real-world applications as illustrated above, it is surprising, then, that there are no existing general-purpose approaches for drawing such comparisons. While there is some previous work on the comparison of document sets (referred to as {\em comparative text mining}), these existing approaches lack the generality to be widely applicable across different use case scenarios with different comparison criteria.  Moreover, much of the work in the area focuses largely on the summarization of shared or unshared topics among document groups (\eg \newcite{Wan2011Summarizing}, \newcite{Huang2011Comparative}, \newcite{Campr2013Topic}, \newcite{Wang2012Comparative}, \newcite{Zhai2004Crosscollection}). That is, the problem of drawing {\em multi-faceted} comparisons among the groups themselves is not typically addressed.  This, then, motivates our development of a {\em general-purpose} model for comparisons of document sets along arbitrary dimensions. We use this model for the identification of similarities, differences, trends, and anomalies among large {\em groups} of documents.  We begin by formally describing our model.

\section{Our~Formal~Model~for Comparing~Document~Groups}
\label{sec:model}

As input, we are given several groups of documents, and our task is to compare them. We now formally define these document groups and the criteria used to compare them.  Let $D =\{d_1, d_2, \ldots,d_N\}$ be a document collection comprising the totality of documents under consideration, where $N$ is the size. Let $D^P$ be a partition of $D$ representing the document groups.   
\begin{defn}
\label{defn:group}
A document group is a subset $D^{P}_i \in D^P$ (where index $i \in \{1 \dots |D^P|\}$).
\end{defn} 

Each document group in $D^P$, for instance, might represent articles associated with either a particular organization (\eg university), a research funding source (\eg NSF or DARPA program), or a time period (\eg a fiscal year).  Document groups are compared using {\em comparison criteria}, $D^C$, a family of subsets of $D$.

\begin{defn}
\label{defn:comparison}
A comparison criterion is a subset $D^{C}_i \in D^C$ (where index $i \in \{1 \dots |D^C|\}$).
\end{defn}

Intuitively, each subset of $D^C$ represents a set of documents sharing some attribute. Our model allows great flexibility in how $D^C$ is defined.  For instance, $D^C$ might be defined by the named entities mentioned within documents (\eg each subset contains documents that mention a particular person or organization of interest).  For the present work, we define $D^C$ by topics discovered using latent Dirichlet allocation or LDA \cite{Blei2003Latent}.   

~\\
\noindent
{\bf LDA Topics as Comparison  Criteria.}  Probabilistic topic modeling algorithms like LDA
discover latent themes (\ie topics) in document collections.  By using these discovered topics as the comparison criteria, we can compare arbitrary groups of documents by the themes and subject areas comprising them.  Let $K$ be the number of topics or themes in $D$.   Each document in $D$ is composed of a sequence of words:  $d_i=\langle s_{i1}, s_{i2}, \ldots, s_{iN_i}\rangle$, where $N_i$ is the number of words in $d_i$ and $i \in \{1 \ldots N\}$.  $V=\bigcup_{i=1}^{N}f(d_i)$ is the vocabulary of $D$, where $f(\cdot)$ takes a sequence of elements and returns a set.  LDA takes $K$ and $D$ (including its components such as $V$) as input and produces two matrices as output, one of which is $\theta$.  The matrix $\theta \in \mathbb{R}^{N \times K}$ is the document-topic distribution matrix and shows the distribution of topics within each document.  Each row of the matrix represents a probability distribution.  $D^C$ is constructed using $K$ subsets of documents, each of which represent a set of documents pertaining largely to the same topic. That is, for $t \in \{1\ldots K\}$ and $i \in \{1 \ldots N\}$, each subset $D^{C}_{t} \in D^C$ is comprised of all documents $d_i$ where $t = \argmax_{x}{\theta_{ix}}$.\footnote{~$D^C$ is also a partition of $D$, when defined in this way.}  Having defined the document groups $D^P$ and the comparison criteria $D^C$, we now construct a bipartite graph model used to perform comparisons.

\begin{figure}[htb]
\begin{center}
\centerline{\fbox{\includegraphics[scale=0.4]{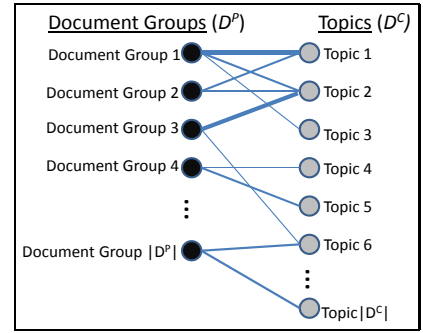}}}
\caption{{\footnotesize {\bf [Toy Illustration of Bipartite Graph Model.]} Each black node (\ie node $\in P$) represents a document group.  Each gray node (\ie node $\in C$) represents a cluster of documents pertaining primarily to the same topic. }}
\label{fig:bipartite}
\end{center}
\vskip -0.2in
\end{figure}

~\\
\noindent
{\bf A Bipartite Graph Model.} Our objective is to compare the {\em document groups} in $D^P$ based on $D^C$.  We do so by representing $D^P$ and $D^C$ as a weighted bipartite graph, $G=(P,C,E,w)$, where $P$ and $C$ are disjoint sets of nodes, $E$ is the edge set, and $w:E\rightarrow\mathbb{Z}^+$ are the edge weights. 
Each subset of $D^P$ is represented as a node in $P$, and each subset of $D^C$ is represented as a node in $C$. Let $\alpha:P\rightarrow D^P$ and $\beta:C\rightarrow D^C$ be functions that map nodes to the document subsets that they represent.  Then, the edge set $E$ is $\{(u,v)\mid u\in P, v\in C, \alpha(u) \cap \beta(v) \neq \emptyset \}$, and the edge weight for any two nodes $u \in P$ and $v \in C$ is $w((u,v)) = |\alpha(u) \cap \beta(v)|$.  Concisely, each weighted edge in G between a document group (in $P$) and a topic (in $C$) represents the number of documents shared among the two sets. Figure \ref{fig:bipartite} shows a toy illustration of the model.  Each node in $P$ is shown in black and represents a subset of $D^P$ (\ie a document group).  Each node in $C$ is shown in gray and represents a subset of $D^C$ (\ie a document cluster pertaining primarily to the same topic).  Each edge represents the intersection of the two subsets it connects.   In the next section, we will describe basic algorithms on such bipartite graphs capable of yielding important insights into the similarities and differences among document groups.  


\section{Basic Algorithms Using the Model}
\label{sec:alg}

We focus on three basic operations in this work.

~\\
\noindent
{\bf Node Entropy.}  Let $\vec{w}$~be a vector of weights for all edges incident to some node $v \in E$.  The {\em entropy}           $\mathrm{H}$ of $v$ is:
$\mathrm{H}(v)= -\sum_{i} p_i\log_{|\vec{w}|}(p_i)$, where $p_i = \frac{w_i}{\sum_j{w_j}}$ and $i,j \in \{1 \ldots |\vec{w}|\}$. A similar formulation was employed in \newcite{Eagle2010Network}.  Intuitively, if $v \in P$, $H(v)$ measures the extent to which the document group is concentrated around a small number of topics (lower values of $H(v)$ mean more concentrated).  Similarly, if $v \in C$, it is the extent to which a topic is concentrated around a small number of document groups.

~\\
\noindent
{\bf Node Similarity.} Given a graph $G$, there are many ways to measure the similarity of two nodes based on their connections.  Such measures can be used to infer similarity (and dissimilarity) among document groups.  However, existing methods are not well-suited for the task of document group comparison. The well-known SimRank algorithm \cite{Jeh2002SimRank} ignores edge weights, and neither SimRank nor its extension, SimRank++~\cite{Antonellis2008Simrank}, scale to larger graphs.  SimRank++ and ASCOS~\cite{Chen2013ASCOS} do incorporate edge weights but in ways that are not appropriate for document group comparisons.  For instance, both SimRank++ and ASCOS incorporate magnitude in the similarity computation.  Consider the case where document groups are defined as research labs.  ASCOS and SimRank++ will measure large research labs and small research labs as less similar when in fact they may publish nearly identical lines of research.  Finally, under these existing methods, document groups sharing zero topics in common could still be considered similar, which is undesirable here.  For these reasons, we formulate similarity as follows.   Let $N^G(\cdot)$ be a function that returns the neighbors of a given node in $G$.  Given two nodes $u,v \in P$, let $L^{u,v} = N^G(u) \cup N^G(v)$ and let $x:I \rightarrow L^{u,v}$ be the indexing function for $L^{u,v}$.\footnote{$I$ is the index set of $L^{u,v}$.}  We construct two vectors, $\vec{a}$ and $\vec{b}$, where  $a_k = w(u,x(k))$, $b_k = w(v,x(k))$, and $k \in I$.  Each vector is essentially a sequence of weights for edges between $u,v \in P$ and each node in $L^{u,v}$.  Similarity of two nodes is measured using the cosine similarity of their corresponding sequences,  $\frac{\vec{a} \cdot \vec{b}}{\|\vec{a}\| \|\vec{b}\|}$, which we compute using a function $sim(\cdot,\cdot)$. Thus, document groups are considered more similar when they have similar sets of topics in similar proportions.  As we will show later, this simple solution, based on item-based collaborative filtering \cite{Sarwar2001Itembased}, is surprisingly effective at inferring similarity among document groups in $G$.

~\\
{\bf Node Clusters.}  Identifying clusters of related nodes in the bipartite graph $G$ can show how document groups form larger classes.  However, we find that $G$ is typically fairly dense.  For these reasons, partitioning of the one-mode projection of $G$ and other standard bipartite graph clustering techniques (\eg \newcite{Dhillion2001Coclustering} and \newcite{Sun2009Rankingbased}) are rendered less effective.  We instead employ a different tack and exploit the node similarities computed earlier. We transform $G$ into a new weighted graph $G^P=(P, E^P, w^{sim})$ where $E^P = \{(u,v) \mid u,v \in P, sim(u,v) > \xi \}$, $\xi$ is a pre-defined threshold, and $w^{sim}$ is the edge weight function (\ie $w^{sim}=sim$).  Thus, $G^P$ is the similarity graph of document groups.  $\xi=0.5$ was used as the threshold for our analyses.  To find clusters in $G^P$, we employ the Louvain algorithm, a heuristic method based on modularity optimization \cite{Blondel2008Fast}.  Modularity measures the fraction of edges falling within clusters as compared to the expected fraction if edges were distributed evenly in the graph \cite{Newman2006Modularity}.  The algorithm initially assigns each node to its own cluster.  At each iteration, in a local and greedy fashion, nodes are re-assigned to clusters with which they achieve the highest modularity.

\section{Example Analysis: NSF Grants}
\label{sec:eval}

As a realistic and informative case study, we utilize our model to characterize funding programs of the National Science Foundation (NSF).  This corpus consists of 132,372 grant abstracts describing awards for basic research and other support funded by the NSF between the years 1990 and 2002~\cite{Bache2013UCI}.\footnote{Data for years 1989 and 2003 in this publicly available corpus were partially missing and omitted in some analyses.}  Each award is associated with both a program element (\ie funding source) and a date.  We define {\em document groups} in two ways: by program element and by calendar year.  For comparison criteria, we used topics discovered with the MALLET implementation of LDA \cite{McCallum2002MALLET} using $K=400$ as the number of topics and $200$ as the number of iterations.  All other parameters were left as defaults.  The NSF corpus possesses unique properties that lend themselves to experimental evaluation. For instance, program elements are not only associated with specific sets of research topics but are named based on the content of the program.  This provides a measure of ground truth against which we can validate our model.  We structure our analyses around specific questions, which now follow.

~\\
\noindent
{\bf Which NSF programs are focused on specific areas and which are not?} When defining {\em document groups} as program elements (\ie each NSF program is a node in $P$), node entropy can be used to answer this question.  Table \ref{tab:entropy.program} shows examples of program elements most and least associated with specific topics, as measured by entropy. For example, the program {\em 1311 Linguistics} (low entropy) is largely focused on a single {\em linguistics} topic (labeled by LDA with words such as ``language,'' ``languages,'' and ``linguistic'').  By contrast, the {\em Australia} program (high entropy) was designed to support US-Australia cooperative research across many fields, as correctly inferred by our model.

\begin{table}[htb]
\centering
{\scriptsize
\begin{tabular}{|l|l|} \hline
\multicolumn{2}{|c|}{{\bf Low Entropy Program Elements}}  \\ \hline
\multicolumn{1}{|c|}{{\bf Program}}   & \multicolumn{1}{|c|}{{\bf Primary LDA Topic}}     \\\hline

{\em 1311 Linguistics}      & language languages linguistic   \\ \hline
{\em 4091 Network Infrastructure}    &  network connection internet         \\ \hline \hline
\multicolumn{2}{|c|}{{\bf High Entropy Program Elements}}  \\ \hline
\multicolumn{1}{|c|}{{\bf Program}}   & \multicolumn{1}{|c|}{{\bf Primary LDA Topic}}     \\\hline
  {\em 5912 Australia}           & (many topics \& disciplines)  \\ \hline
{\em 9130 Res. Improvements in Minority Instit.}     & (many topics \& disciplines)                    \\ 

     \hline
\end{tabular}
\caption{{\footnotesize {\bf [Examples of High/Low Entropy Programs.]}  }}
\label{tab:entropy.program}
}
\vskip -0.1in
\end{table}

\begin{figure}[htb]
\begin{center}
\centerline{\fbox{\includegraphics[scale=0.25]{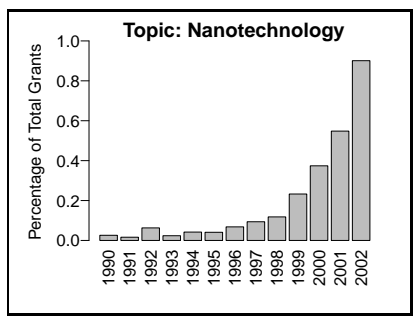}}}
\caption{{\footnotesize {\bf [Uptrend in Nanotechnology.]}  Our model correctly identifies the surge in nanotechnology R\&D beginning in the late 1990s.}}
\label{fig:yearplot}
\end{center}
\vskip -0.4in
\end{figure}

~\\
\noindent
{\bf Which research areas are growing/emerging?} When defining {\em document groups} as calendar years (instead of program elements), low entropy nodes in $C$ are topics concentrated around certain years. Concentrations in later years indicate growth. The LDA-discovered topic {\em nanotechnology} is among the lowest entropy topics (\ie an outlier topic with respect to entropy).  As shown in Figure \ref{fig:yearplot}, the number of {\em nanotechnology} grants drastically increased in proportion through 2002.  This result is consistent with history, as the National Nanotechnology Initiative was proposed in the late 1990s to promote nanotechnology R\&D.\footnote{See {\em National Nanotechnology Initiative} at \\\url{http://en.wikipedia.org} for more information.} One could also measure such trends using budget allocations by incorporating the award amounts into the edge weights of $G$.

\begin{figure}[htb]
\begin{center}
\centerline{\fbox{\includegraphics[scale=0.25]{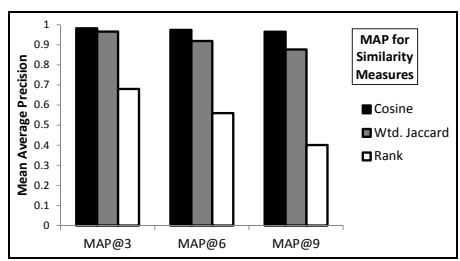}}}
\caption{{\footnotesize {\bf [Mean Average Precision (MAP).]}  Cosine similarity outperforms alternative approaches.}}
\label{fig:perfmap}
\end{center}
\vskip -0.4in
\end{figure}

~\\
\noindent
{\bf Given an NSF program, to which other programs is it most similar?}  As described in Section \ref{sec:alg}, when each node in $P$ represents an NSF program, our model can easily identify the programs most similar to a given program.  For instance, Table \ref{tab:sim} shows the top three most similar programs to both the {\em Theoretical Physics} and {\em Ecology} programs. Results agree with intuition. For each NSF program, we identified the top $n$ most similar programs ranked by our $sim(\cdot, \cdot)$ function, where $n \in \{3, 6, 9\}$. These programs were manually judged for relatedness, and the Mean Average Precision (MAP), a standard performance metric for ranking tasks in information retrieval, was computed. We were unsuccessful in evaluating alternative weighted similarity measures mentioned in Section \ref{sec:alg} due to their aforementioned issues with scalability and the size of the NSF dataset. (For instance, the implementations of ASCOS~\cite{Antonellis2008Simrank} and SimRank~\cite{Jeh2002SimRank} that we considered are available here.\footnote{See \texttt{networkx\_addon} project at \\\url{http://github.com/hhchen1105/}.})  Recall that our $sim(\cdot, \cdot)$ function is based on measuring the cosine similarity between two weight vectors, $\vec{a}$ and $\vec{b}$, generated from our bipartite graph model.   As a baseline for comparison, we evaluated two additional similarity implementations using these weight vectors.  The first measures the similarity between weight vectors using weighted Jaccard similarity, which is   $\frac{\sum_k \min (a_k, b_k)}{\sum_k \max (a_k, b_k)}$ (denoted as {\em Wtd. Jaccard}).  The second measure is implemented by taking the Spearman's rank correlation coefficient of $\vec{a}$ and $\vec{b}$ (denoted as {\em Rank}). Figure \ref{fig:perfmap} shows the Mean Average Precision (MAP) for each method and each value of $n$. With the exception of the difference between  {\em Cosine} and {\em Wtd. Jaccard} for MAP$@3$, all other performance differentials were statistically significant, based on a one-way ANOVA and post-hoc Tukey HSD at a 5\% significance level. This, then, provides some validation for our choice.

\begin{table}[htb]
\centering
{\scriptsize
\begin{tabular}{|l|l|} \hline
{\bf 1245 Theoretical Physics}           & {\bf 1182 Ecology}            \\ \hline  \hline

1286 Elementary Particle Theory     & 1128 Ecological Studies    \\ \hline
1287 Mathematical Physics    &  1196 Environmental Biology         \\ \hline
1284 Atomic Theory             &   1195 Ecological Research \\ \hline
\end{tabular}
\caption{{\footnotesize {\bf [Similarity Queries.]}  Three most similar programs to the {\em Theoretical Physics} and {\em Ecology} programs.}}
\label{tab:sim}
}
\vskip -0.1in
\end{table}

\begin{figure}[htb]
\begin{center}
\centerline{\fbox{\includegraphics[scale=0.12]{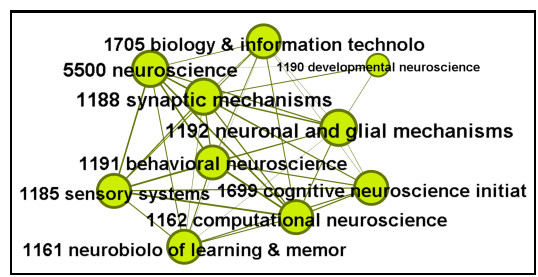}}}
\caption{{\footnotesize {\bf [Neuroscience Programs.]} A discovered cluster of program elements all related to {\em neuroscience}.}}
\label{fig:neuroscience}
\end{center}
\vskip -0.2in
\end{figure}

~\\
\noindent
{\bf How do NSF programs join together to form larger program categories?}  As mentioned, by using the similarity graph $G^P$ constructed from $G$, clusters of related NSF programs can be discovered.  Figure \ref{fig:neuroscience}, for instance, shows a discovered cluster of NSF programs all related to the field of neuroscience. Each NSF program (\ie node) is composed of many documents.  

~\\
\noindent
{\bf Which pairs of grants are the most similar in the research they describe?} Although the focus of this paper is on drawing comparisons among {\em groups} of documents, it is often necessary to draw comparisons among {\em individual} documents, as well.  For instance, in the case of this NSF corpus, one may wish to identify pairs of grants from different programs describing highly similar lines of research.  One common approach to this is to exploit the low-dimensional representations of documents returned by LDA \cite{Blei2003Latent}.  Any given document $d_i \in D$ (where $i \in \{1 \ldots N\}$) can be represented by a K-dimensional probability vector of topic proportions given by $\theta_{i*}$, the $i^{th}$ row of the document-topic matrix $\theta$. The similarity between any two documents, then, can be measured using the distance between their corresponding probability vectors (\ie probability distributions). We quantify the similarity between probability vectors using the complement of Hellinger distance:  $H_S(d_x,d_y) = 1-\frac{1}{\sqrt{2}}\sqrt{\sum^{K}_{i=1}(\sqrt{\theta_{xi}}-\sqrt{\theta_{yi}})^2 }$, where $x,y \in \{1 \ldots N\}$.  Unfortunately, identifying the set of {\em most} similar document pairs in this way can be computationally expensive, as the number of pairwise comparisons scales quadratically with the size of the corpus.  For the moderately-sized NSF corpus, this amounts to well over 8 billion comparisons. To address this issue, our bipartite graph model can be exploited as a {\em blocking} heuristic using either the document groups or the comparison criteria.  In the latter case, one can limit the pairwise comparisons to only those documents that reside in the same subset of $D^C$.  For the former case, {\em node similarity} can be used.  Instead of comparing each document with every other document, we can limit the comparisons to only those document groups of interest that are deemed similar by our model.  As an illustrative example, out of the $665$ different NSF programs covering these $132,372$ grant abstracts, the program {\em 1271 Computational Mathematics} and the program {\em 2865 Numeric, Symbolic, and Geometric Computation} are inferred as being highly similar by our model.  Thus, we can limit the pairwise comparisons to only such {\em document groups} that are similar and likely to {\em contain} similar documents.  In the case of these two programs, the following two grants are easily identified as being the most similar with a Hellinger similarity ($H_S$) score of $0.73$ (only text snippets are shown due to space constraints):
\balance

\begin{quote}
{\footnotesize
\ul{Grant \#1}  \\
{\em Program}:  1271 Computational Mathematics\\\\
{\em Title}:  Analyses of Structured {\bf Computational Problems} and {\bf Parallel} Iterative {\bf Algorithms}.   \\\\
{\em Abstract}:  The main objectives of the research planned is the analysis of {\bf large scale} structured {\bf computational problems} and of the convergence of {\bf parallel} iterative methods for solving {\bf linear systems} and applications of these techniques to the solution of large {\bf sparse} and dense structured systems of {\bf linear equations}
}
\end{quote}

\begin{quote}
{\footnotesize
\ul{Grant \#2}  \\
{\em Program}:  2865 Numeric, Symbolic, and Geometric Computation\\\\
{\em Title}:  {\bf Sparse} {\bf Matrix} {\bf Algorithms} on {\bf Distributed} Memory Multiprocessors.  \\\\
{\em Abstract}:  The design, analysis, and implementation of {\bf algorithms} for the solution of {\bf sparse} {\bf matrix} problems on {\bf distributed} memory multiprocessors will be investigated. The development of these {\bf parallel} {\bf sparse} {\bf matrix} {\bf algorithms} should have an impact of challenging {\bf large-scale} {\bf computational problems} in several scientific, econometric, and engineering disciplines.
}
\end{quote}

\noindent
Some key terms in each grant are manually highlighted in bold.  As can be seen, despite some differences in terminology, the two lines of research are related, as matrices (studied in Grant \#2) are used to compactly represent and work with systems of linear equations (studied in Grant \#1).  That is, despite such differences in terminology (\eg ``matrix'' vs. ``linear systems'', ``parallel'' vs. ``distributed''), document similarity can still be accurately inferred by taking the Hellinger similarity of the LDA-derived low-dimensional representations for the two documents.  In this way, by exploiting the {\em group-level} similarities inferred by our model in combination with such document-level similarities, we can more effectively ``zero in'' on such highly related document pairs.

\section{Conclusion}

We have presented a bipartite graph model for drawing comparisons among large {\em groups} of documents.  We showed how basic algorithms using the model can identify trends and anomalies among the document groups. As an example analysis, we demonstrated how our model can be used to better characterize and evaluate NSF research programs. For future work, we plan on employing alternative comparison criteria in our model such as those derived from named entity recognition and paraphrase detection.

\clearpage
\newpage
\balance

\end{document}